\documentclass[11pt,a4paper]{article}
\usepackage[hyperref]{acl2021}
\usepackage{times}
\usepackage{latexsym}

\usepackage{booktabs}
\usepackage{graphicx}
\usepackage[normalem]{ulem}
\useunder{\uline}{\ul}{}
\usepackage{adjustbox} 
\usepackage{bbold}     
\usepackage{bm}        
\usepackage{amsmath}
\usepackage{dirtytalk} 
\usepackage{lscape}    
\usepackage{mathtools} 
\usepackage{tabularx}

\usepackage{microtype}

\aclfinalcopy 
\def\aclpaperid{118} 

\title{DeCLUTR: Deep Contrastive Learning for Unsupervised Textual Representations}

\author{%
  \(\text{John~Giorgi}^{1,5,6}\; \text{Osvald~Nitski}^{2,7}\;
  \text{Bo~Wang}^{1,4,6,7,\dag}\; \text{Gary~Bader}^{1,3,5,\dag}\;\) \\
  \(^{1} \text{Department of Computer Science, University of Toronto}\)\\
  \(^{2} \text{Faculty of Applied Science and Engineering, University of Toronto}\) \\
  \(^{3} \text{Department of Molecular Genetics, University of Toronto}\) \\
  \(^{4} \text{Department of Laboratory Medicine and Pathobiology, University of Toronto}\) \\
  \(^{5} \text{Terrence Donnelly Centre for Cellular \& Biomolecular Research}\) \\
  \(^{6} \text{Vector Institute for Artificial Intelligence}\) \\
  \(^{7} \text{Peter Munk Cardiac Center, University Health Network}\) \\
  \(^{\dag} \text{Co-senior authors}\) \\
  \texttt{\{john.giorgi, osvald.nitski, gary.bader\}@mail.utoronto.ca} \\
  \texttt{bowang@vectorinstitute.ai}\\
}

\date{}

\begin{document}
\maketitle
\begin{abstract}
Sentence embeddings are an important component of many natural language processing (NLP) systems. Like word embeddings, sentence embeddings are typically learned on large text corpora and then transferred to various downstream tasks, such as clustering and retrieval. Unlike word embeddings, the highest performing solutions for learning sentence embeddings require labelled data, limiting their usefulness to languages and domains where labelled data is abundant. In this paper, we present DeCLUTR: Deep Contrastive Learning for Unsupervised Textual Representations. Inspired by recent advances in deep metric learning (DML), we carefully design a self-supervised objective for learning universal sentence embeddings that does not require labelled training data. When used to extend the pretraining of transformer-based language models, our approach closes the performance gap between unsupervised and supervised pretraining for universal sentence encoders. Importantly, our experiments suggest that the quality of the learned embeddings scale with both the number of trainable parameters and the amount of unlabelled training data. Our code and pretrained models are publicly available and can be easily adapted to new domains or used to embed unseen text.\footnote{\url{https://github.com/JohnGiorgi/DeCLUTR}}
\end{abstract}

\section{Introduction}

Due to the limited amount of labelled training data available for many natural language processing (NLP) tasks, transfer learning has become ubiquitous \citep{ruder-etal-2019-transfer}. For some time, transfer learning in NLP was limited to pretrained \textit{word} embeddings \citep{word2vec, glove}. Recent work has demonstrated strong transfer task performance using pretrained \textit{sentence} embeddings. These fixed-length vectors, often referred to as \say{universal} sentence embeddings, are typically learned on large corpora and then transferred to various downstream tasks, such as clustering (e.g. topic modelling) and retrieval (e.g. semantic search). Indeed, sentence embeddings have become an area of focus, and many supervised \citep{infersent}, semi-supervised \citep{gensen, STILTS, cer-etal-2018-universal, sentence-bert} and unsupervised \citep{doc2vec, discsent, skip-thought, fastsent, quick-thoughts} approaches have been proposed. However, the highest performing solutions require labelled data, limiting their usefulness to languages and domains where labelled data is abundant. Therefore, closing the performance gap between unsupervised and supervised universal sentence embedding methods is an important goal.

Pretraining transformer-based language models has become the primary method for learning textual representations from unlabelled corpora \citep{radford2018improving, BERT, Transformer-XL, XLNet, roberta, ELECTRA}. This success has primarily been driven by masked language modelling (MLM). This self-supervised \textit{token}-level objective requires the model to predict the identity of some randomly masked tokens from the input sequence. In addition to MLM, some of these models have mechanisms for learning \textit{sentence}-level embeddings via self-supervision. In BERT \citep{BERT}, a special classification token is prepended to every input sequence, and its representation is used in a binary classification task to predict whether one textual segment follows another in the training corpus, denoted Next Sentence Prediction (NSP). However, recent work has called into question the effectiveness of NSP \citep{lample2019cross, you2019reducing, SpanBERT}. In RoBERTa \citep{roberta}, the authors demonstrated that removing NSP during pretraining leads to unchanged or even slightly improved performance on downstream sentence-level tasks (including semantic text similarity and natural language inference). In ALBERT \citep{ALBERT}, the authors hypothesize that NSP conflates \textit{topic} prediction and \textit{coherence} prediction, and instead propose a Sentence-Order Prediction objective (SOP), suggesting that it better models inter-sentence coherence. In preliminary evaluations, we found that neither objective produces good universal sentence embeddings (see \autoref{pretrained-transformers-make-poor-sentence-encoders}). Thus, we propose a simple but effective self-supervised, sentence-level objective inspired by recent advances in metric learning.

Metric learning is a type of representation learning that aims to learn an embedding space where the vector representations of similar data are mapped close together, and vice versa \citep{lowe1995similarity, mika1999fisher, xing2003distance}. In computer vision (CV), deep metric learning (DML) has been widely used for learning visual representations \citep{wohlhart2015learning, wen2016discriminative, zhang2016zero, bucher2016improving, leal2016learning, tao2016siamese, hermans2017defense, he2018triplet, grabner20183d, yelamarthi2018zero, yu2018hard}. Generally speaking, DML is approached as follows: a \say{pretext} task (often self-supervised, e.g. colourization or inpainting) is carefully designed and used to train deep neural networks to generate useful feature representations. Here, \say{useful} means a representation that is easily adaptable to other downstream tasks, unknown at training time. Downstream tasks (e.g. object recognition) are then used to evaluate the quality of the learned features (independent of the model that produced them), often by training a linear classifier on the task using these features as input. The most successful approach to date has been to design a pretext task for learning with a pair-based contrastive loss function. For a given \textit{anchor} data point, contrastive losses attempt to make the distance between the anchor and some \textit{positive} data points (those that are similar) smaller than the distance between the anchor and some \textit{negative} data points (those that are dissimilar) \citep{contrastive}. The highest-performing methods generate anchor-positive pairs by randomly augmenting the same image (e.g. using crops, flips and colour distortions); anchor-negative pairs are randomly chosen, augmented views of different images \citep{viewsMI, tian2019contrastive, moco, simclr}. In fact, \citealp{kong2019mutual} demonstrate that the MLM and NSP objectives are also instances of contrastive learning.

Inspired by this approach, we propose a self-supervised, contrastive objective that can be used to pretrain a sentence encoder. Our objective learns universal sentence embeddings by training an encoder to minimize the distance between the embeddings of textual segments randomly sampled from nearby in the same document. We demonstrate our objective's effectiveness by using it to extend the pretraining of a transformer-based language model and obtain state-of-the-art results on SentEval \citep{senteval} -- a benchmark of 28 tasks designed to evaluate universal sentence embeddings. Our primary contributions are:

\begin{itemize}
    \item We propose a self-supervised sentence-level objective that
    can be used alongside MLM to pretrain transformer-based language models, inducing generalized embeddings for sentence- and paragraph-length text without any labelled data (\autoref{comparison-to-baselines}).
    \item We perform extensive ablations to determine which factors are important for learning high-quality embeddings (\autoref{ablation}).
    \item We demonstrate that the quality of the learned embeddings scale with model and data size. Therefore, performance can likely be improved simply by collecting more unlabelled text or using a larger encoder (\autoref{scalability}).
    \item We open-source our solution and provide detailed instructions for training it on new data or embedding unseen text.\footnote{\url{https://github.com/JohnGiorgi/DeCLUTR}}
\end{itemize}

\section{Related Work} \label{universal-sentence-embeddings}

Previous works on universal sentence embeddings can be broadly grouped by whether or not they use labelled data in their pretraining step(s), which we refer to simply as \textit{supervised or semi-supervised} and \textit{unsupervised}, respectively.

\paragraph{Supervised or semi-supervised}
The highest performing universal sentence encoders are pretrained on the human-labelled natural language inference (NLI) datasets Stanford NLI (SNLI) \citep{SNLI} and MultiNLI \citep{MultiNLI}. NLI is the task of classifying a pair of sentences (denoted the \say{hypothesis} and the \say{premise}) into one of three relationships: \textit{entailment}, \textit{contradiction} or \textit{neutral}. The effectiveness of NLI for training universal sentence encoders was demonstrated by the supervised method InferSent \citep{infersent}. Universal Sentence Encoder (USE) \citep{cer-etal-2018-universal} is semi-supervised, augmenting an unsupervised, Skip-Thoughts-like task (\citealt{skip-thought}, see \autoref{unsupervised}) with supervised training on the SNLI corpus. The recently published Sentence Transformers \citep{sentence-bert} method fine-tunes pretrained, transformer-based language models like BERT \citep{BERT} using labelled NLI datasets.

\paragraph{Unsupervised} \label{unsupervised}
Skip-Thoughts \citep{skip-thought} and FastSent \citep{fastsent} are popular unsupervised techniques that learn sentence embeddings by using an encoding of a sentence to predict words in neighbouring sentences. However, in addition to being computationally expensive, this generative objective forces the model to reconstruct the \textit{surface} form of a sentence, which may capture information irrelevant to the \textit{meaning} of a sentence. QuickThoughts \citep{quick-thoughts} addresses these shortcomings with a simple discriminative objective; given a sentence and its context (adjacent sentences), it learns sentence representations by training a classifier to distinguish context sentences from non-context sentences. The unifying theme of unsupervised approaches is that they exploit the \say{distributional hypothesis}, namely that the meaning of a word (and by extension, a sentence) is characterized by the word context in which it appears.

\begin{figure}[t]
\centering
\includegraphics[width=\linewidth]{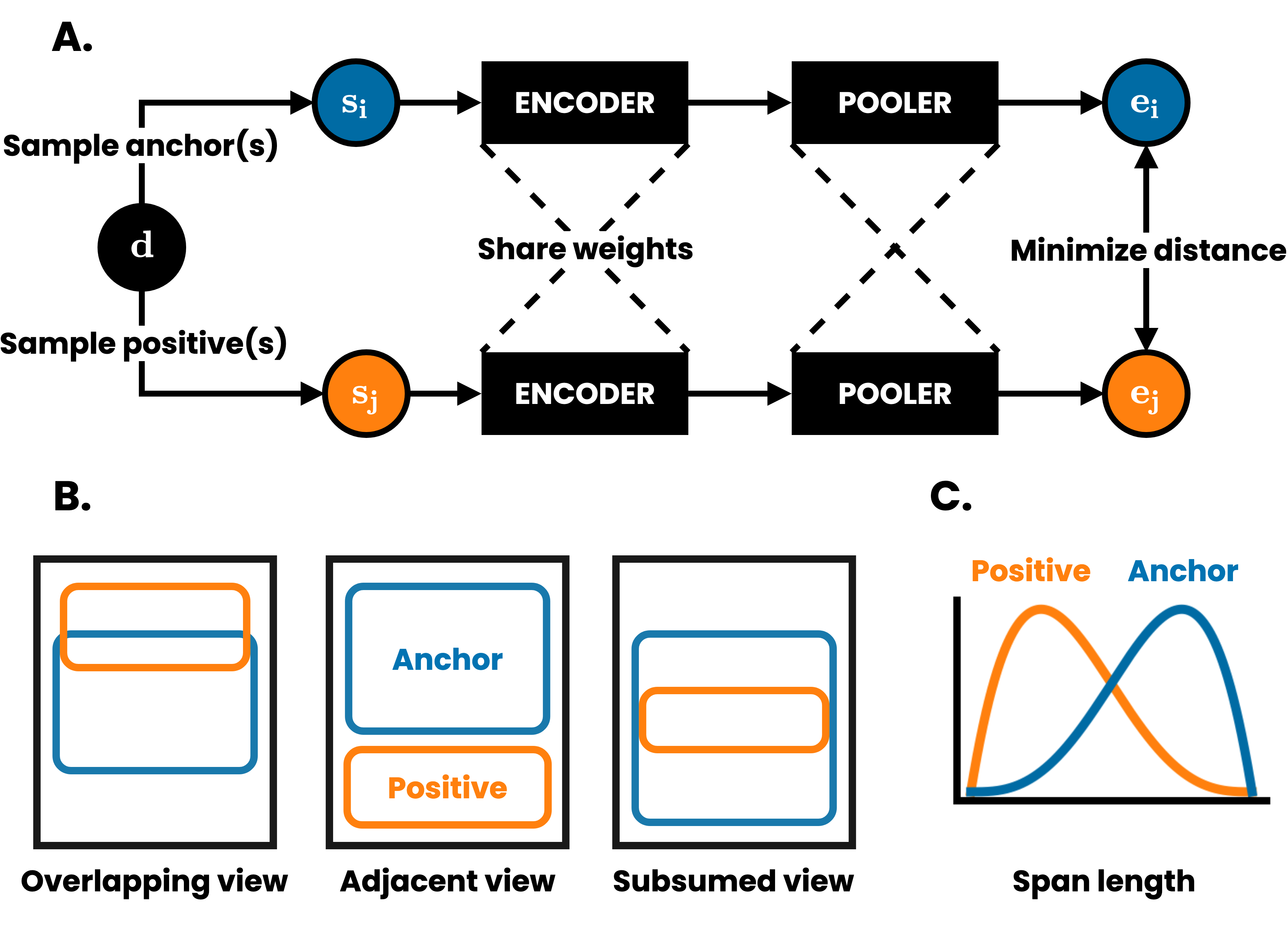}
\caption{Overview of the self-supervised contrastive objective. (A) For each document \(d\) in a minibatch of size \(N\), we sample \(A\) anchor spans per document and \(P\) positive spans per anchor. For simplicity, we illustrate the case where \(A=P=1\) and denote the anchor-positive span pair as \(\bm{s}_i, \bm{s}_j\). Both spans are fed through the same encoder \(f(\cdot)\) and pooler \(g(\cdot)\) to produce the corresponding embeddings \(\bm{e}_i = g(f(\bm{s}_i)), \bm{e}_{j} = g(f(\bm{s}_{j}))\). The encoder and pooler are trained to minimize the distance between embeddings via a contrastive prediction task, where the other embeddings in a minibatch are treated as negatives (omitted here for simplicity). (B) Positive spans can overlap with, be adjacent to or be subsumed by the sampled anchor span. (C) The length of anchors and positives are randomly sampled from beta distributions, skewed toward longer and shorter spans, respectively.}
\label{fig:01}
\end{figure}

\medskip
Our overall approach is most similar to Sentence Transformers -- we extend the pretraining of a transformer-based language model to produce useful sentence embeddings -- but our proposed objective is self-supervised. Removing the dependence on labelled data allows us to exploit the vast amount of unlabelled text on the web without being restricted to languages or domains where labelled data is plentiful (e.g. English Wikipedia). Our objective most closely resembles QuickThoughts; some distinctions include: we relax our sampling to textual \textit{segments} of up to paragraph length (rather than natural \textit{sentences}), we sample one or more positive segments per anchor (rather than strictly one), and we allow these segments to be adjacent, overlapping or subsuming (rather than strictly adjacent; see \autoref{fig:01}, B).

\section{Model} \label{model}

\subsection{Self-supervised contrastive loss}

Our method learns textual representations via a contrastive loss by maximizing agreement between textual segments (referred to as \say{spans} in the rest of the paper) sampled from nearby in the same document. Illustrated in \autoref{fig:01}, this approach comprises the following components:

\begin{itemize}
    \item A data loading step randomly samples paired anchor-positive spans from each document in a minibatch of size \(N\). Let \(A\) be the number of anchor spans sampled per document, \(P\) be the number of positive spans sampled per anchor and \(i \in \{1 \dots AN\}\) be the index of an arbitrary anchor span. We denote an anchor span and its corresponding \(p \in \{1 \dots P\}\) positive spans as \(\bm{s}_i\) and \(\bm{s}_{i + pAN}\) respectively. This procedure is designed to maximize the chance of sampling semantically similar anchor-positive pairs (see \autoref{span-sampling}). 
    \item An encoder \(f(\cdot)\) maps each token in the input spans to an embedding.  Although our method places no constraints on the choice of encoder, we chose \(f(\cdot)\) to be a transformer-based language model, as this represents the state-of-the-art for text encoders (see \autoref{continued-pretraining}).
    \item A pooler \(g(\cdot)\)  maps the encoded spans \(f(\bm{s}_i)\) and \(f(\bm{s}_{i + pAN})\) to fixed-length embeddings \(\bm{e}_i = g(f(\bm{s}_i))\) and its corresponding mean positive embedding 
    
    {
    \begin{align*}
        \bm{e}_{i + AN} = \frac{1}{P}\sum_{p=1}^{P} g(f(\bm{s}_{i + pAN}))
    \end{align*}}
    
    Similar to \citealt{sentence-bert}, we found that choosing \(g(\cdot)\) to be the mean of the token-level embeddings (referred to as \say{mean pooling} in the rest of the paper) performs well (see Appendix, \autoref{tab:s1}). We pair each anchor embedding with the mean of multiple positive embeddings. This strategy was proposed by \citealt{arora2019theoretical}, who demonstrated theoretical and empirical improvements compared to using a single positive example for each anchor.
    \item A contrastive loss function defined for a contrastive prediction task. Given a set of embedded spans \(\{\bm{e}_k\}\) including a positive pair of examples \(\bm{e}_i\) and \(\bm{e}_{i + AN}\), the contrastive prediction task aims to identify \(\bm{e}_{i + AN}\) in \(\{\bm{e}_k\}_{k\not=i}\) for a given \(\bm{e}_i\)
    
    {\small
    \begin{align*}
        \ell(i, j) &= -\log\frac{\exp(\text{sim}(\bm{e}_i, \bm{e}_{j}) / \tau)}{\sum_{k=1}^{2AN} \mathbb{1}_{[i \not = k]} \cdot \exp(\text{sim}(\bm{e}_i, \bm{e}_k) / \tau)} 
    \end{align*}}
    
    where \(\text{sim}(\bm{u}, \bm{v}) = \bm{u}^T\bm{v} / ||\bm{u}||_2||\bm{v}||_2\) denotes the cosine similarity of two vectors \(\bm{u}\) and \(\bm{v}\), \(\mathbb{1}_{[i \not = k]} \in \{0, 1\}\) is an indicator function evaluating to 1 if \(i \not = k\), and \(\tau > 0\) denotes the temperature hyperparameter.
\end{itemize}

During training, we randomly sample minibatches of \(N\) documents from the train set and define the contrastive prediction task on anchor-positive pairs \(\bm{e}_i, \bm{e}_{i + AN}\) derived from the \(N\) documents, resulting in \(2AN\) data points. As proposed in \citep{sohn2016improved}, we treat the other \(2(AN - 1)\) instances within a minibatch as negative examples. The cost function takes the following form

{
\begin{align*}
    \mathcal{L}_{\text{contrastive}} &= \sum_{i=1}^{AN} \ell(i, i + AN) + \ell(i + AN, i) 
\end{align*}}

\noindent  This is the InfoNCE loss used in previous works \citep{sohn2016improved, wu2018unsupervised, oord2018representation} and denoted normalized temperature-scale cross-entropy loss or \say{NT-Xent} in \citep{simclr}. To embed text with a trained model, we simply pass batches of tokenized text through the model, without sampling spans. Therefore, the computational cost of our method at test time is the cost of the encoder, \(f(\cdot)\), plus the cost of the pooler, \(g(\cdot)\), which is negligible when using mean pooling.

\subsection{Span sampling} \label{span-sampling}

We start by choosing a minimum and maximum span length; in this paper, \(\ell_{\text{min}} = 32\) and \(\ell_{\text{max}} = 512\), the maximum input size for many pretrained transformers. Next, a document \(d\) is tokenized to produce a sequence of \(n\) tokens \(\bm{x}^d = (x_1, x_2 \dots x_n)\). To sample an anchor span \(\bm{s}_i\) from \(\bm{x}^d\), we first sample its length \(\ell_{\text{anchor}}\) from a beta distribution and then randomly (uniformly) sample its starting position \(s_i^{\text{start}}\)

{
\begin{align*}
\ell_{\text{anchor}} &= \big\lfloor p_{\text{anchor}} \times (\ell_{\text{max}} - \ell_{\text{min}}) + \ell_{\text{min}}\big\rfloor \\
s_i^{\text{start}} &\sim \{0, \dots, n - \ell_{\text{anchor}}\} \\
s_i^{\text{end}} &= s_i^{\text{start}} + \ell_{\text{anchor}} \\
\bm{s}_i &= \bm{x}^d_{s_i^{\text{start}}:s_i^{\text{end}}}
\end{align*}
}

\noindent We then sample \(p \in \{1 \dots P\}\) corresponding positive spans \(\bm{s}_{i + pAN}\) independently following a similar procedure

{
\begin{align*}
\ell_{\text{positive}} &= \big\lfloor p_{\text{positive}} \times (\ell_{\text{max}} - \ell_{\text{min}}) + \ell_{\text{min}} \big\rfloor \\
 s_{i + pAN}^{\text{start}} &\sim \{s_i^{\text{start}} - \ell_{\text{positive}}, \dots, s_i^{\text{end}}\} \\
s_{i + pAN}^{\text{end}} &= s_{i + pAN}^{\text{start}} + \ell_{\text{positive}} \\
\bm{s}_{i + pAN} &= \bm{x}^d_{s_{i + pAN}^{\text{start}}:s_{i + pAN}^{\text{end}}}
\end{align*}
}

\noindent where \(p_{\text{anchor}} \sim \text{Beta}(\alpha=4, \beta=2)\), which skews anchor sampling towards longer spans, and \(p_{\text{positive}} \sim \text{Beta}(\alpha=2, \beta=4)\), which skews positive sampling towards shorter spans (\autoref{fig:01}, C). In practice, we restrict the sampling of anchor spans from the same document such that they are a minimum of \(2 * \ell_{\text{max}}\) tokens apart. In \autoref{span-examples}, we show examples of text that has been sampled by our method. We note several carefully considered decisions in the design of our sampling procedure:

\begin{itemize}
    \item Sampling span lengths from a distribution clipped at \(\ell_{\text{min}}=32\) and \(\ell_{\text{max}}=512\) encourages the model to produce good embeddings for text ranging from sentence- to paragraph-length. At test time, we expect our model to be able to embed up-to paragraph-length texts.
    \item We found that sampling longer lengths for the anchor span than the positive spans improves performance in downstream tasks (we did not find performance to be sensitive to the specific choice of \(\alpha\) and \(\beta\)). The rationale for this is twofold. First, it enables the model to learn global-to-local view prediction as in \citep{deepMI, viewsMI, simclr} (referred to as \say{subsumed view} in \autoref{fig:01}, B). Second, when \(P > 1\), it encourages diversity among positives spans by lowering the amount of repeated text.
    \item Sampling positives nearby to the anchor exploits the distributional hypothesis and increases the chances of sampling valid (i.e. semantically similar) anchor-positive pairs. 
    \item By sampling multiple anchors per document, each anchor-positive pair is contrasted against both \textit{easy} negatives (anchors and positives sampled from \textit{other} documents in a minibatch) and \textit{hard} negatives (anchors and positives sampled from the \textit{same} document).
\end{itemize}

\noindent In conclusion, the sampling procedure produces three types of positives: positives that partially overlap with the anchor, positives adjacent to the anchor, and positives subsumed by the anchor (\autoref{fig:01}, B) and two types of negatives: easy negatives sampled from a different document than the anchor, and hard negatives sampled from the same document as the anchor. Thus, our stochastically generated training set and contrastive loss implicitly define a family of predictive tasks which can be used to train a model, independent of any specific encoder architecture.

\subsection{Continued MLM pretraining} \label{continued-pretraining}

We use our objective to extend the pretraining of a transformer-based language model \citep{attention-is-all-you-need}, as this represents the state-of-the-art encoder in NLP. We implement the MLM objective as described in \citep{BERT} on each anchor span in a minibatch and sum the losses from the MLM and contrastive objectives before backpropagating

\begin{align*}
    \mathcal{L} &= \mathcal{L}_{\text{contrastive}} + \mathcal{L}_{\text{MLM}}
\end{align*}

\noindent This is similar to existing pretraining strategies, where an MLM loss is paired with a sentence-level loss such as NSP \citep{BERT} or SOP \citep{ALBERT}. To make the computational requirements feasible, we do not train from scratch, but rather we continue training a model that has been pretrained with the MLM objective. Specifically, we use both RoBERTa-base \citep{roberta} and DistilRoBERTa \citep{distilbert} (a distilled version of RoBERTa-base) in our experiments. In the rest of the paper, we refer to our method as DeCLUTR-small (when extending DistilRoBERTa pretraining) and DeCLUTR-base (when extending RoBERTa-base pretraining).

\section{Experimental setup}

\subsection{Dataset, training, and implementation}

\paragraph{Dataset} We collected all documents with a minimum token length of 2048 from OpenWebText \citep{openwebtext} an open-access subset of the WebText corpus \citep{radford2019language}, yielding 497,868 documents in total. For reference, Google's USE was trained on 570,000 human-labelled sentence pairs from the SNLI dataset (among other unlabelled datasets). InferSent and Sentence Transformer models were trained on both SNLI and MultiNLI, a total of 1 million human-labelled sentence pairs.

\paragraph{Implementation} We implemented our model in PyTorch \citep{pytorch} using AllenNLP \citep{AllenNLP}. We used the NT-Xent loss function implemented by the PyTorch Metric Learning library \citep{pytorch-metric-learning} and the pretrained transformer architecture and weights from the Transformers library \citep{hf-transformers}. All models were trained on up to four NVIDIA Tesla V100 16 or 32GB GPUs.

\paragraph{Training} \label{training} Unless specified otherwise, we train for one to three epochs over the 497,868 documents with a minibatch size of 16 and a temperature \(\tau = 5 \times 10^{-2}\) using the AdamW optimizer \citep{adamw} with a learning rate (LR) of \(5 \times 10^{-5}\) and a weight decay of \(0.1\). For every document in a minibatch, we sample two anchor spans (\(A=2\)) and two positive spans per anchor (\(P=2\)). We use the Slanted Triangular LR scheduler \citep{ULMFiT} with a number of train steps equal to training instances and a cut fraction of \(0.1\). The remaining hyperparameters of the underlying pretrained transformer (i.e. DistilRoBERTa or RoBERTa-base) are left at their defaults. All gradients are scaled to a vector norm of \(1.0\) before backpropagating. Hyperparameters were tuned on the SentEval validation sets.

\begin{table}[t]
\centering
\caption{Trainable model parameter counts and sentence embedding dimensions. DeCLUTR-small and DeCLUTR-base are pretrained DistilRoBERTa and RoBERTa-base models respectively after continued pretraining with our method.}
\label{tab:01}
\renewcommand{\arraystretch}{0.90}
\resizebox{\linewidth}{!}{%
\begin{tabular}{@{}lcc@{}}
\toprule
Model                        & Parameters  & Embedding dim. \\ \midrule
\multicolumn{3}{c}{\textit{Bag-of-words (BoW) baselines}}   \\ \midrule
GloVe                        & --       & 300               \\
fastText                     & --       & 300               \\ \midrule
\multicolumn{3}{c}{\textit{Supervised and semi-supervised}} \\ \midrule
InferSent                    & 38M      & 4096              \\
Universal Sentence Encoder   & 147M     & 512               \\
Sentence Transformers        & 125M     & 768               \\ \midrule
\multicolumn{3}{c}{\textit{Unsupervised}}                   \\ \midrule
QuickThoughts                & 73M      & 4800              \\
DeCLUTR-small                & 82M      & 768               \\
DeCLUTR-base                 & 125M     & 768               \\ \bottomrule
\end{tabular}%
}
\end{table}

\begin{table*}[t]
\centering
\caption{Results on the downstream tasks from the test set of SentEval. QuickThoughts scores are taken directly from \citep{quick-thoughts}. USE: Google's Universal Sentence Encoder. Transformer-small and Transformer-base are pretrained DistilRoBERTa and RoBERTa-base models respectively, using mean pooling. DeCLUTR-small and DeCLUTR-base are pretrained DistilRoBERTa and RoBERTa-base models respectively after continued pretraining with our method. Average scores across all tasks, excluding SNLI, are shown in the top half of the table. Bold: best scores. \(\Delta\): difference to DeCLUTR-base average score. \(\color{green} \uparrow\) and \(\color{red} \downarrow\) denote increased or decreased performance with respect to the underlying pretrained model. *: Unsupervised evaluations.}
\label{tab:02}
\renewcommand{\arraystretch}{0.90}
\resizebox{\textwidth}{!}{%
\begin{tabular}{@{}llllllllllll@{}}
\toprule
Model &
  \multicolumn{1}{c}{CR} &
  \multicolumn{1}{c}{MR} &
  \multicolumn{1}{c}{MPQA} &
  \multicolumn{1}{c}{SUBJ} &
  \multicolumn{1}{c}{SST2} &
  \multicolumn{1}{c}{SST5} &
  \multicolumn{1}{c}{TREC} &
  \multicolumn{1}{c}{MRPC} &
  \multicolumn{1}{c}{SNLI} &
  Avg. &
  \(\Delta\) \\ \midrule
\multicolumn{12}{c}{\textit{Bag-of-words (BoW) weak baselines}} \\ \midrule
GloVe &
  78.78 &
  77.70 &
  87.76 &
  91.25 &
  80.29 &
  44.48 &
  83.00 &
  73.39/81.45 &
  65.85 &
  \multicolumn{1}{l}{65.47} &
  \multicolumn{1}{l}{-13.63} \\
fastText &
  79.18 &
  78.45 &
  87.88 &
  91.53 &
  82.15 &
  45.16 &
  83.60 &
  74.49/82.44 &
  68.79 &
  \multicolumn{1}{l}{68.56} &
  \multicolumn{1}{l}{-10.54} \\ \midrule
\multicolumn{12}{c}{\textit{Supervised and semi-supervised}} \\ \midrule
InferSent &
  84.37 &
  79.42 &
  89.04 &
  93.03 &
  84.24 &
  45.34 &
  90.80 &
  76.35/83.48 &
  84.16 &
  \multicolumn{1}{l}{76.00} &
  \multicolumn{1}{l}{-3.10} \\
USE &
  85.70 &
  79.38 &
  88.89 &
  93.11 &
  84.90 &
  46.11 &
  \textbf{95.00} &
  72.41/82.01 &
  83.25 &
  \multicolumn{1}{l}{78.89} &
  \multicolumn{1}{l}{-0.21} \\
Sent. Transformers &
  \textbf{90.78} &
  84.98 &
  88.72 &
  92.67 &
  \textbf{90.55} &
  \textbf{52.76} &
  87.40 &
  76.64/82.99 &
  \textbf{84.18} &
  \multicolumn{1}{l}{77.19} &
  \multicolumn{1}{l}{-1.91} \\ \midrule
\multicolumn{12}{c}{\textit{Unsupervised}} \\ \midrule
QuickThoughts &
  86.00 &
  82.40 &
  \textbf{90.20} &
  94.80 &
  87.60 &
  \multicolumn{1}{c}{--} &
  92.40 &
  \textbf{76.90/84.00} &
  \multicolumn{1}{c}{--} &
  -- &
  -- \\
Transformer-small &
  86.60 &
  82.12 &
  87.04 &
  94.77 &
  88.03 &
  49.50 &
  91.60 &
  74.55/81.75 &
  71.88 &
  \multicolumn{1}{l}{72.58} &
  \multicolumn{1}{l}{-6.52} \\
Transformer-base &
  88.19 &
  84.35 &
  86.49 &
  95.28 &
  89.46 &
  51.27 &
  93.20 &
  74.20/81.44 &
  72.19 &
  \multicolumn{1}{l}{72.70} &
  \multicolumn{1}{l}{-6.40} \\
DeCLUTR-small &
  87.52 \(\color{green} \uparrow\) &
  82.79 \(\color{green} \uparrow\) &
  87.87 \(\color{green} \uparrow\) &
  94.96 \(\color{green} \uparrow\) &
  87.64 \(\color{red} \downarrow\) &
  48.42 \(\color{red} \downarrow\) &
  90.80 \(\color{red} \downarrow\) &
  75.36/82.70 \(\color{green} \uparrow\) &
  73.59 \(\color{green} \uparrow\) &
  \multicolumn{1}{l}{77.50 \(\color{green} \uparrow\)} &
  \multicolumn{1}{l}{-1.60} \\
DeCLUTR-base &
  90.68 \(\color{green} \uparrow\) &
  \textbf{85.16 \(\color{green} \uparrow\)} &
  88.52 \(\color{green} \uparrow\) &
  \textbf{95.78 \(\color{green} \uparrow\)} &
  90.01 \(\color{green} \uparrow\) &
  51.18 \(\color{red} \downarrow\) &
  93.20 \(\color{green} \uparrow\) &
  74.61/82.65 \(\color{green} \uparrow\) &
  74.74 \(\color{green} \uparrow\) &
  \multicolumn{1}{l}{\textbf{79.10 \(\color{green} \uparrow\)}} &
  -- \\ \midrule
Model &
  \multicolumn{1}{c}{SICK-E} &
  \multicolumn{1}{c}{SICK-R} &
  \multicolumn{1}{c}{STS-B} &
  \multicolumn{1}{c}{COCO} &
  \multicolumn{1}{c}{STS12*} &
  \multicolumn{1}{c}{STS13*} &
  \multicolumn{1}{c}{STS14*} &
  \multicolumn{1}{c}{STS15*} &
  \multicolumn{1}{c}{STS16*} &
   &
   \\ \midrule
GloVe &
  78.89 &
  72.30 &
  62.86 &
  0.40 &
  53.44 &
  51.24 &
  55.71 &
  59.62 &
  57.93 &
  -- &
  -- \\
fastText &
  79.01 &
  72.98 &
  68.26 &
  0.40 &
  58.85 &
  58.83 &
  63.42 &
  69.05 &
  68.24 &
  -- &
  -- \\
InferSent &
  \textbf{86.30} &
  \textbf{83.06} &
  78.48 &
  \textbf{65.84} &
  62.90 &
  56.08 &
  66.36 &
  74.01 &
  72.89 &
  -- &
  -- \\
USE &
  85.37 &
  81.53 &
  \textbf{81.50} &
  62.42 &
  \textbf{68.87} &
  71.70 &
  \textbf{72.76} &
  \textbf{83.88} &
  \textbf{82.78} &
  -- &
  -- \\
Sent. Transformers &
  82.97 &
  79.17 &
  74.28 &
  60.96 &
  64.10 &
  65.63 &
  69.80 &
  74.71 &
  72.85 &
  -- &
  -- \\
QuickThoughts &
  \multicolumn{1}{c}{--} &
  \multicolumn{1}{c}{--} &
  \multicolumn{1}{c}{--} &
  60.55 &
  \multicolumn{1}{c}{--} &
  \multicolumn{1}{c}{--} &
  \multicolumn{1}{c}{--} &
  \multicolumn{1}{c}{--} &
  \multicolumn{1}{c}{--} &
  -- &
  -- \\
Transformer-small &
  81.96 &
  77.51 &
  70.31 &
  60.48 &
  53.99 &
  45.53 &
  57.23 &
  65.57 &
  63.51 &
  -- &
  -- \\
Transformer-base &
  80.29 &
  76.84 &
  69.62 &
  60.14 &
  53.28 &
  46.10 &
  56.17 &
  64.69 &
  62.79 &
  -- &
  -- \\
DeCLUTR-small &
  83.46 \(\color{green} \uparrow\) &
  77.66 \(\color{green} \uparrow\) &
  77.51 \(\color{green} \uparrow\) &
  60.85 \(\color{green} \uparrow\) &
  63.66 \(\color{green} \uparrow\) &
  68.93 \(\color{green} \uparrow\) &
  70.40 \(\color{green} \uparrow\) &
  78.25 \(\color{green} \uparrow\) &
  77.74 \(\color{green} \uparrow\) &
  -- &
  -- \\
DeCLUTR-base &
  83.84 \(\color{green} \uparrow\) &
  78.62 \(\color{green} \uparrow\) &
  79.39 \(\color{green} \uparrow\) &
  62.35 \(\color{green} \uparrow\) &
  63.56 \(\color{green} \uparrow\) &
  \textbf{72.58 \(\color{green} \uparrow\)} &
  71.70 \(\color{green} \uparrow\) &
  79.95 \(\color{green} \uparrow\) &
  79.59 \(\color{green} \uparrow\) &
  -- &
  -- \\ \bottomrule
\end{tabular}%
}
\end{table*}

\begin{table*}[!ht]
\centering
\caption{Results on the probing tasks from the test set of SentEval. USE: Google's Universal Sentence Encoder. Transformer-small and Transformer-base are pretrained DistilRoBERTa and RoBERTa-base models respectively, using mean pooling. DeCLUTR-small and DeCLUTR-base are pretrained DistilRoBERTa and RoBERTa-base models respectively after continued pretraining with our method. Bold: best scores. \(\color{green} \uparrow\) and \(\color{red} \downarrow\) denote increased or decreased performance with respect to the underlying pretrained model.}
\label{tab:03}
\renewcommand{\arraystretch}{0.90}
\resizebox{\textwidth}{!}{%
\begin{tabular}{@{}llllllllllll@{}}
\toprule
Model &
  SentLen &
  WC &
  TreeDepth &
  TopConst &
  BShift &
  Tense &
  SubjNum &
  ObjNum &
  SOMO &
  CoordInv &
  Avg. \\ \midrule
\multicolumn{12}{c}{\textit{Bag-of-words (BoW) weak baselines}} \\ \midrule
GloVe &
  57.82 &
  81.10 &
  31.41 &
  62.70 &
  49.74 &
  83.58 &
  78.39 &
  76.31 &
  49.55 &
  53.62 &
  62.42 \\
fastText &
  55.46 &
  82.10 &
  32.74 &
  63.32 &
  50.16 &
  86.68 &
  79.75 &
  79.81 &
  50.21 &
  51.41 &
  63.16 \\ \midrule
\multicolumn{12}{c}{\textit{Supervised and semi-supervised}} \\ \midrule
InferSent &
  78.76 &
  \textbf{89.50} &
  37.72 &
  \textbf{80.16} &
  61.41 &
  88.56 &
  \textbf{86.83} &
  83.91 &
  52.11 &
  66.88 &
  72.58 \\
USE &
  73.14 &
  69.44 &
  30.87 &
  73.27 &
  58.88 &
  83.81 &
  80.34 &
  79.14 &
  56.97 &
  61.13 &
  66.70 \\
Sent. Transformers &
  69.21 &
  51.79 &
  30.08 &
  50.38 &
  69.70 &
  83.02 &
  79.74 &
  77.85 &
  60.10 &
  60.33 &
  63.22 \\ \midrule
\multicolumn{12}{c}{\textit{Unsupervised}} \\ \midrule
Transformer-small &
  88.62 &
  65.00 &
  \textbf{40.87} &
  75.38 &
  88.63 &
  87.84 &
  86.68 &
  84.17 &
  63.75 &
  64.78 &
  74.57 \\
Transformer-base &
  81.96 &
  59.67 &
  38.84 &
  74.02 &
  \textbf{90.08} &
  88.59 &
  85.51 &
  83.33 &
  \textbf{68.54} &
  \textbf{71.32} &
  74.19 \\
DeCLUTR-small (ours) &
  \textbf{88.85 \(\color{green} \uparrow\)} &
  74.87 \(\color{green} \uparrow\) &
  38.48 \(\color{red} \downarrow\) &
  75.17 \(\color{red} \downarrow\) &
  86.12 \(\color{red} \downarrow\) &
  88.71 \(\color{green} \uparrow\) &
  86.31 \(\color{red} \downarrow\) &
  \textbf{84.30 \(\color{green} \uparrow\)} &
  61.27 \(\color{red} \downarrow\) &
  62.98 \(\color{red} \downarrow\) &
  \textbf{74.71 \(\color{green} \uparrow\)} \\
DeCLUTR-base (ours) &
  84.62 \(\color{green} \uparrow\) &
  68.98 \(\color{green} \uparrow\) &
  38.35 \(\color{red} \downarrow\) &
  74.78 \(\color{green} \uparrow\) &
  87.85 \(\color{red} \downarrow\) &
  \textbf{88.82 \(\color{green} \uparrow\)} &
  86.56 \(\color{green} \uparrow\) &
  83.88 \(\color{green} \uparrow\) &
  65.08 \(\color{red} \downarrow\) &
  67.54 \(\color{red} \downarrow\) &
  74.65 \(\color{green} \uparrow\) \\ \bottomrule
\end{tabular}%
}
\end{table*}

\subsection{Evaluation}

We evaluate all methods on the SentEval benchmark, a widely-used toolkit for evaluating general-purpose, fixed-length sentence representations. SentEval is divided into 18 \textit{downstream} tasks -- representative NLP tasks such as sentiment analysis, natural language inference, paraphrase detection and image-caption retrieval -- and ten \textit{probing} tasks, which are designed to evaluate what linguistic properties are encoded in a sentence representation. We report scores obtained by our model and the relevant baselines on the downstream and probing tasks using the SentEval toolkit\footnote{\url{https://github.com/facebookresearch/SentEval}} with default parameters (see \autoref{senteval-details} for details). Note that all the supervised approaches we compare to are trained on the SNLI corpus, which is included as a downstream task in SentEval. To avoid train-test contamination, we compute average downstream scores without considering SNLI when comparing to these approaches in \autoref{tab:02}.

\subsubsection{Baselines}

We compare to the highest performing, most popular sentence embedding methods: InferSent, Google's USE and Sentence Transformers. For InferSent, we compare to the latest model.\footnote{\url{https://dl.fbaipublicfiles.com/infersent/infersent2.pkl}} We use the latest \say{large} USE model\footnote{\url{https://tfhub.dev/google/universal-sentence-encoder-large/5}}, as it is most similar in terms of architecture and number of parameters to DeCLUTR-base. For Sentence Transformers, we compare to \say{roberta-base-nli-mean-tokens}\footnote{\url{https://www.sbert.net/docs/pretrained_models.html}}, which, like DeCLUTR-base, uses the RoBERTa-base architecture and pretrained weights. The only difference is each method's extended pretraining strategy. We include the performance of averaged GloVe\footnote{\url{http://nlp.stanford.edu/data/glove.840B.300d.zip}} and fastText\footnote{\url{https://dl.fbaipublicfiles.com/fasttext/vectors-english/crawl-300d-2M.vec.zip}} word vectors as weak baselines. Trainable model parameter counts and sentence embedding dimensions are listed in \autoref{tab:01}. Despite our best efforts, we could not evaluate the pretrained QuickThought models against the full SentEval benchmark. We cite the scores from the paper directly. Finally, we evaluate the pretrained transformer model's performance before it is subjected to training with our contrastive objective, denoted \say{Transformer-*}. We use mean pooling on the pretrained transformers token-level output to produce sentence embeddings -- the same pooling strategy used in our method. 

\section{Results} \label{results}

In \autoref{comparison-to-baselines}, we compare the performance of our model against the relevant baselines. In the remaining sections, we explore which components contribute to the quality of the learned embeddings.

\subsection{Comparison to baselines} \label{comparison-to-baselines}

\paragraph{Downstream task performance}
Compared to the underlying pretrained models DistilRoBERTa and RoBERTa-base, 
DeCLUTR-small and DeCLUTR-base obtain large boosts in average downstream performance, +4\% and +6\% respectively (\autoref{tab:02}). DeCLUTR-base leads to improved or equivalent performance for every downstream task but one (SST5) and DeCLUTR-small for all but three (SST2, SST5 and TREC). Compared to existing methods, DeCLUTR-base matches or even outperforms average performance without using any hand-labelled training data. Surprisingly, we also find that DeCLUTR-small outperforms Sentence Transformers while using \(\sim\)34\% less trainable parameters. 

\paragraph{Probing task performance}
With the exception of InferSent, existing methods perform poorly on the probing tasks of SentEval (\autoref{tab:03}). Sentence Transformers, which begins with a pretrained transformer model and fine-tunes it on NLI datasets, scores approximately 10\% lower on the probing tasks than the model it fine-tunes. In contrast, both DeCLUTR-small and DeCLUTR-base perform comparably to the underlying pretrained model in terms of average performance. We note that the purpose of the probing tasks is \textit{not} the development of ad-hoc models that attain top performance on them \citep{probing}. However, it is still interesting to note that high downstream task performance can be obtained without sacrificing probing task performance. Furthermore, these results suggest that fine-tuning transformer-based language models on NLI datasets may discard some of the linguistic information captured by the pretrained model's weights. We suspect that the inclusion of MLM in our training objective is responsible for DeCLUTR's relatively high performance on the probing tasks.

\paragraph{Supervised vs. unsupervised downstream tasks}
The downstream evaluation of SentEval includes supervised and unsupervised tasks. In the unsupervised tasks, the embeddings of the method to evaluate are used as-is without any further training (see \autoref{senteval-details} for details). Interestingly, we find that USE performs particularly well across the unsupervised evaluations in SentEval (tasks marked with a * in \autoref{tab:02}). Given the similarity of the USE architecture to Sentence Transformers and DeCLUTR and the similarity of its supervised NLI training objective to InferSent and Sentence Transformers, we suspect the most likely cause is one or more of its additional training objectives. These include a conversational response prediction task \citep{henderson2017efficient} and a Skip-Thoughts \citep{skip-thought} like task.

\begin{figure}[t]
\centering
\includegraphics[width=\linewidth]{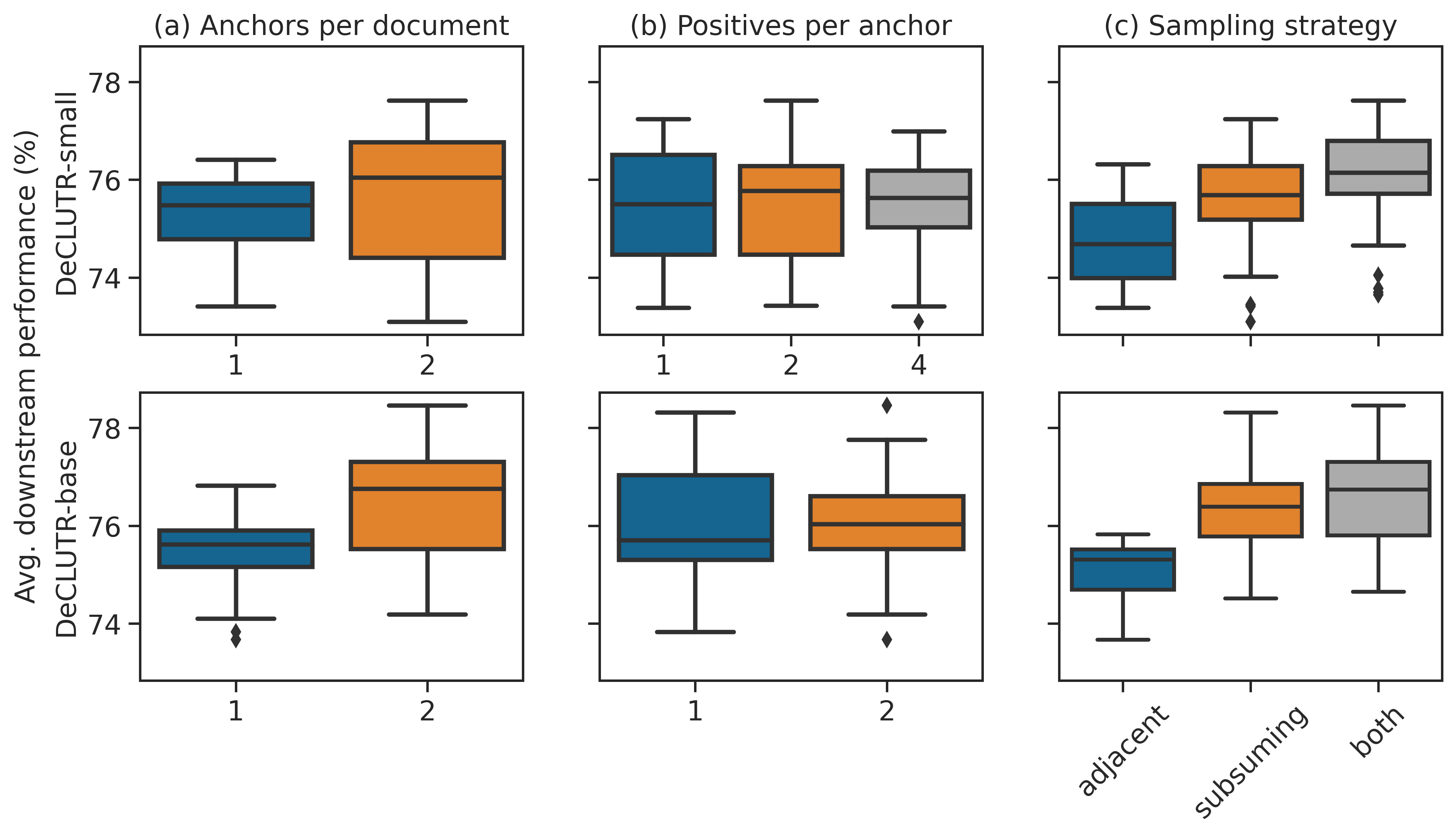}
\caption{Effect of the number of anchor spans sampled per document (a), the number of positive spans sampled per anchor (b), and the sampling strategy (c). Averaged downstream task scores are reported from the validation set of SentEval. Performance is computed over a grid of hyperparameters and plotted as a distribution. The grid is defined by all permutations of number of anchors \(A=\{1, 2\}\), number of positives \(P=\{1, 2, 4\}\), temperatures \(\tau = \{5 \times 10^{-3}, 1 \times 10^{-2}, 5 \times 10^{-2}\}\) and learning rates \(\alpha = \{5 \times 10^{-5}, 1 \times 10^{-4}\}\). \(P=4\) is omitted for DeCLUTR-base as these experiments did not fit into GPU memory.}
\label{fig:02}
\end{figure}

\begin{figure}[t]
\centering
\includegraphics[width=\linewidth]{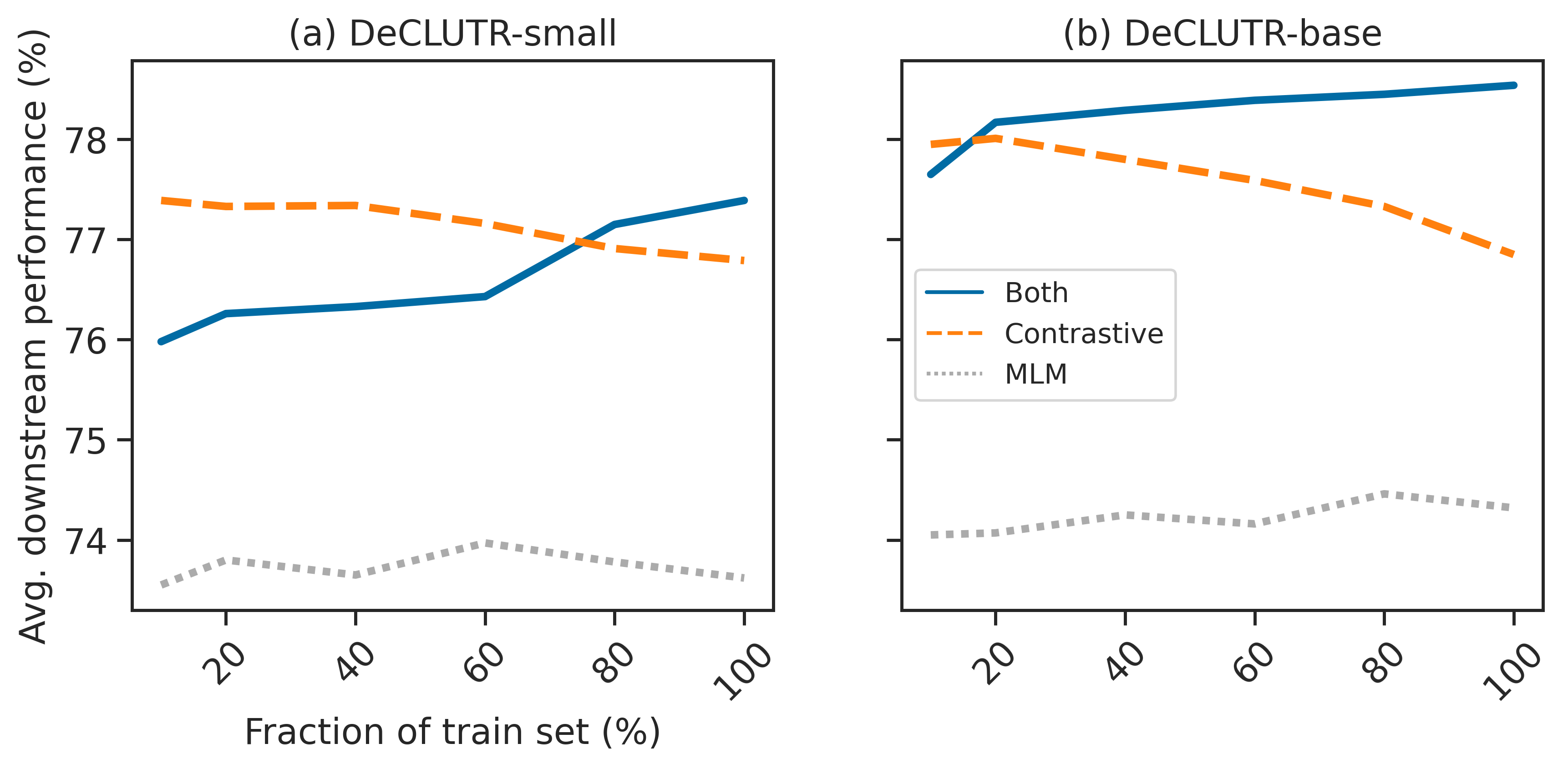}
\caption{Effect of training objective, train set size and model capacity on SentEval performance. DeCLUTR-small has 6 layers and ${\sim}$82M parameters. DeCLUTR-base has 12 layers and ${\sim}$125M parameters. Averaged downstream task scores are reported from the validation set of SentEval. 100\% corresponds to 1 epoch of training with all 497,868 documents from our OpenWebText subset.}
\label{fig:03}
\end{figure}

\subsection{Ablation of the sampling procedure} \label{ablation}

We ablate several components of the sampling procedure, including the number of anchors sampled per document \(A\), the number of positives sampled per anchor \(P\), and the sampling strategy for those positives (\autoref{fig:02}). We note that when \(A=2\), the model is trained on twice the number of spans and twice the effective batch size (\(2AN\), where \(N\) is the number of documents in a minibatch) as compared to when \(A=1\). To control for this, all experiments where \(A=1\) are trained for two epochs (twice the number of epochs as when \(A=2\)) and for two times the minibatch size (\(2N\)). Thus, both sets of experiments are trained on the same number of spans and the same effective batch size (\(4N\)), and the only difference is the number of anchors sampled per document (\(A\)).

We find that sampling multiple anchors per document has a large positive impact on the quality of learned embeddings. We hypothesize this is because the difficulty of the contrastive objective increases when \(A > 1\). Recall that a minibatch is composed of random documents, and each anchor-positive pair sampled from a document is contrasted against all other anchor-positive pairs in the minibatch. When \(A > 1\), anchor-positive pairs will be contrasted against other anchors and positives from the same document, increasing the difficulty of the contrastive objective, thus leading to better representations. We also find that a positive sampling strategy that allows positives to be adjacent to \textit{and} subsumed by the anchor outperforms a strategy that only allows adjacent \textit{or} subsuming views, suggesting that the information captured by these views is complementary. Finally, we note that sampling multiple positives per anchor (\(P > 1\)) has minimal impact on performance. This is in contrast to \citep{arora2019theoretical}, who found both theoretical and empirical improvements when multiple positives are averaged and paired with a given anchor.

\subsection{Training objective, train set size and model capacity} \label{scalability}

To determine the importance of the training objectives, train set size, and model capacity, we trained two sizes of the model with 10\% to 100\% (1 full epoch) of the train set (\autoref{fig:03}). Pretraining the model with both the MLM and contrastive objectives improves performance over training with either objective alone. Including MLM alongside the contrastive objective leads to monotonic improvement as the train set size is increased. We hypothesize that including the MLM loss acts as a form of regularization, preventing the weights of the pretrained model (which itself was trained with an MLM loss) from diverging too dramatically, a phenomenon known as \say{catastrophic forgetting} \citep{mccloskey1989catastrophic, ratcliff1990connectionist}. These results suggest that the quality of embeddings learned by our approach scale in terms of model capacity and train set size; because the training method is completely self-supervised, scaling the train set would simply involve collecting more unlabelled text.

\section{Discussion and conclusion}

In this paper, we proposed a self-supervised objective for learning universal sentence embeddings. Our objective does not require labelled training data and is applicable to any text encoder. We demonstrated the effectiveness of our objective by evaluating the learned embeddings on the SentEval benchmark, which contains a total of 28 tasks designed to evaluate the transferability and linguistic properties of sentence representations. When used to extend the pretraining of a transformer-based language model, our self-supervised objective closes the performance gap with existing methods that require human-labelled training data. Our experiments suggest that the learned embeddings' quality can be further improved by increasing the model and train set size. Together, these results demonstrate the effectiveness and feasibility of replacing hand-labelled data with carefully designed self-supervised objectives for learning universal sentence embeddings. We release our model and code publicly in the hopes that it will be extended to new domains and non-English languages.

\section*{Acknowledgments}

This research was enabled in part by support provided by Compute Ontario (https://computeontario.ca/), Compute Canada (www.computecanada.ca) and the CIFAR AI Chairs Program and partially funded by the US National Institutes of Health (NIH) [U41 HG006623, U41 HG003751).

\bibliography{acl2021}
\bibliographystyle{acl_natbib}

\appendix

\begin{table*}[t]
\small
\centering
\caption{Results on the downstream and probing tasks from the validation set of SentEval. We compare models trained with the Next Sentence Prediction (NSP) and Sentence-Order Prediction (SOP) losses to a model trained with neither, using two different pooling strategies: "*-CLS", where the special classification token is used as its sentence representation, and "*-mean", where each sentence is represented by the mean of its token embeddings.}
\label{tab:s1}
\renewcommand{\arraystretch}{0.90}
\begin{tabular*}{\linewidth}{@{\extracolsep{\fill}}lcccc@{\extracolsep{\fill}}}
\toprule
                    &            &             & \multicolumn{2}{c}{SentEval}    \\ \cmidrule(l){4-5} 
Model               & Parameters & Embed. dim. & Downstream     & Probing        \\ \midrule
\multicolumn{5}{c}{\textit{Bag-of-Words (BoW) weak baselines}}                   \\ \midrule
GloVe               & --         & 300         & 66.05          & 62.93          \\
fastText            & --         & 300         & 68.75          & 63.46          \\ \midrule
\multicolumn{5}{c}{\textit{Trained with Next Sentence Prediction (NSP) loss}}    \\ \midrule
BERT-base-CLS       & 110M       & 768         & 63.53          & 69.57          \\
BERT-base-mean      & 110M       & 768         & 71.98          & 73.37          \\ \midrule
\multicolumn{5}{c}{\textit{Trained with Sentence-Order Prediction (SOP) loss}}   \\ \midrule
ALBERT-base-V2-CLS  & 11M        & 768         & 58.75          & 69.88          \\
ALBERT-base-V2-mean & 11M        & 768         & 69.39          & \textbf{74.83} \\ \midrule
\multicolumn{5}{c}{\textit{Trained with neither NSP or SOP losses}}              \\ \midrule
RoBERTa-base-CLS    & 125M       & 768         & 68.53          & 66.92          \\
RoBERTa-base-mean   & 125M       & 768         & \textbf{72.84} & 74.59          \\ \bottomrule
\end{tabular*}
\end{table*}

\section{Pretrained transformers make poor universal sentence encoders} \label{pretrained-transformers-make-poor-sentence-encoders}

Certain pretrained transformers, such as BERT and ALBERT, have mechanisms for learning sequence-level embeddings via self-supervision. These models prepend every input sequence with a special classification token (e.g. \say{[CLS]}), and its representation is learned using a simple classification task, such as Next Sentence Prediction (NSP) or Sentence-Order Prediction (SOP) (see \citealt{BERT} and \citealt{ALBERT} respectively for details on these tasks). However, during preliminary experiments, we noticed that these models are not good universal sentence encoders, as measured by their performance on the SentEval benchmark \citep{senteval}. As a simple experiment, we evaluated three pretrained transformer models on SentEval: one trained with the NSP loss (BERT), one trained with the SOP loss (ALBERT) and one trained with neither, RoBERTa \citep{roberta}. We did not find that the CLS embeddings produced by models trained against the NSP or SOP losses to outperform that of a model trained without either loss and sometimes failed to outperform a bag-of-words (BoW) baseline (\autoref{tab:s1}). Furthermore, we find that pooling token embeddings via averaging (referred to as \say{mean pooling} in our paper) outperforms pooling via the CLS classification token. Our results are corroborated by \citealt{roberta}, who find that removing NSP loss leads to the same or better results on downstream tasks and \citealt{sentence-bert}, who find that directly using the output of BERT as sentence embeddings leads to poor performances on the semantic similarity tasks of SentEval.

\section{Examples of sampled spans} \label{span-examples}

In \autoref{tab:s2}, we present examples of anchor-positive and anchor-negative pairs generated by our sampling procedure. We show one example for each possible view of a sampled positive, e.g. positives adjacent to, overlapping with, or subsumed by the anchor. For each anchor-positive pair, we show examples of both a \textit{hard} negative (derived from the same document) and an \textit{easy} negative (derived from another document). Recall that a minibatch is composed of random documents, and each anchor-positive pair sampled from a document is contrasted against all other anchor-positive pairs in the minibatch. Thus, hard negatives, as we have described them here, are generated only when sampling multiple anchors per document (\(A > 1\)).

\section{SentEval evaluation details} \label{senteval-details}

SentEval is a benchmark for evaluating the quality of fixed-length sentence embeddings. It is divided into 18 \textit{downstream} tasks, and 10 \textit{probing} tasks. Sentence embedding methods are evaluated on these tasks via a simple interface\footnote{\url{https://github.com/facebookresearch/SentEval}}, which standardizes training, evaluation and hyperparameters. For most tasks, the method to evaluate is used to produce fix-length sentence embeddings, and a simple logistic regression (LR) or multi-layer perception (MLP) model is trained on the task using these embeddings as input. For other tasks (namely several semantic text similarity tasks), the embeddings are used as-is without any further training. Note that this setup is different from evaluations on the popular GLUE benchmark \citep{glue}, which typically use the task data to fine-tune the parameters of the sentence embedding model.

In \autoref{senteval-tasks}, we present the individual tasks of the SentEval benchmark. In \autoref{senteval-average}, we explain our method for computing the average downstream and average probing scores presented in our paper.

\subsection{SentEval tasks} \label{senteval-tasks}

The downstream tasks of SentEval are representative NLP tasks used to evaluate the transferability of fixed-length sentence embeddings. We give a brief overview of the broad categories that divide the tasks below (see \citealt{senteval} for more details):

\begin{itemize}
    \item \textbf{Binary and multi-class classification}: These tasks cover various types of sentence classification, including sentiment analysis (MR \citealt{MR}, SST2 and SST5 \citealt{SST}), question-type (TREC) \citep{TREC}, product reviews (CR) \citep{CR}, subjectivity/objectivity (SUBJ) \citep{SUBJ} and opinion polarity (MPQA) \citep{MPQA}.
    \item \textbf{Entailment and semantic relatedness}: These tasks cover multiple entailment datasets (also known as natural language inference or NLI), including SICK-E \citep{SICK} and the Stanford NLI dataset (SNLI) \citep{SNLI} as well as multiple semantic relatedness datasets including SICK-R and STS-B \citep{STS-B}.
    \item \textbf{Semantic textual similarity} These tasks (STS12 \citealt{STS12}, STS13 \citealt{STS13}, STS14 \citealt{STS14}, STS15 \citealt{STS15} and STS16 \citealt{STS16}) are similar to the semantic relatedness tasks, except the embeddings produced by the encoder are used as-is in a cosine similarity to determine the semantic similarity of two sentences. No additional model is trained on top of the encoder's output.
    \item \textbf{Paraphrase detection} Evaluated on the Microsoft Research Paraphrase Corpus (MRPC) \citep{MRPC}, this binary classification task is comprised of human-labelled sentence pairs, annotated according to whether they capture a paraphrase/semantic equivalence relationship.
\end{itemize}

\begin{landscape}
\begin{table}[t]
\small
\centering
\caption{Examples of text spans generated by our method. During training, we randomly sample one or more anchors from every document in a minibatch. For each anchor, we randomly sample one or more positives adjacent to, overlapping with, or subsumed by the anchor. All anchor-positive pairs are contrasted with every other anchor-positive pair in the minibatch. This leads to \textit{easy} negatives (anchors and positives sampled from \textit{other} documents in a minibatch) and \textit{hard} negatives (anchors and positives sampled from the \textit{same} document). Here, examples are capped at a maximum length of 64 tokens. During training, we sample spans up to a length of 512 tokens.}
\label{tab:s2}
\renewcommand{\arraystretch}{1.75}
\begin{tabularx}{\linewidth}{X X X X}
\toprule
\multicolumn{1}{l}{Anchor} & Positive & Hard negative & Easy negative
\\ \midrule
\multicolumn{4}{c}{\textit{Overlapping view}}
\\ \midrule
immigrant-rights advocates and law enforcement professionals were skeptical of the new program. Any effort by local cops to enforce immigration laws, they felt, would be bad for community policing, since immigrant victims or witnesses of crime wouldn't feel comfortable talking to police. & feel comfortable talking to police. Some were skeptical that ICE's intentions were really to protect public safety, rather than simply to deport unauthorized immigrants more easily. & liberal parts of the country with large immigrant populations, like Santa Clara County in California and Cook County in Illinois, agreed with the critics of Secure Communities. They worried that implementing the program would strain their relationships with immigrant residents. & that a new location is now available for exploration. A good area, in my view, feels like a natural progression of a game world it doesn't seem tacked on or arbitrary. That in turn needs it to relate
\\ \midrule
\multicolumn{4}{c}{\textit{Adjacent view}}
\\ \midrule
if the ash stops belching out of the volcano then, after a few days, the problem will have cleared, so that's one of the factors. "The other is the wind speed and direction." At the moment the weather patterns are very volatile which is what is making it quite difficult, unlike last year, to predict & where the ash will go. "The public can be absolutely confident that airlines are only able to operate when it is safe to do so." Ryanair said it could not see any ash cloud & A British Airways jumbo jet was grounded in Canada on Sunday following fears the engines had been contaminated with volcanic ash & events are processed in FIFO order. When this nextTickQueue is emptied, the event loop considers all operations to have been completed for the current phase and transitions to the next phase.
\\ \midrule
\multicolumn{4}{c}{\textit{Subsumed view}}
\\ \midrule
Far Cry Primal is an action-adventure video game developed by Ubisoft Montreal and published by Ubisoft. It was released worldwide for PlayStation 4 and Xbox One on February 23, 2016, and for Microsoft Windows on March 1, 2016. The game is a spin-off of the main Far Cry series. It is the first Far Cry game set in the Mesolithic Age. & by Ubisoft. It was released worldwide for PlayStation 4 and Xbox One on February 23, 2016, and for Microsoft Windows on March 1, 2016. The game is a spin-off of the main Far Cry series. & Players take on the role of a Wenja tribesman named Takkar, who is stranded in Oros with no weapons after his hunting party is ambushed by a Saber-tooth Tiger. & to such feelings. Fawkes cried out and flew ahead, and Albus Dumbledore followed. Further along the Dementors' path, people were still alive to be fought for. And no matter how much he himself was hurting, while there were still people who needed him he would go on. For \\
\bottomrule
\end{tabularx}
\end{table}
\end{landscape}

\begin{itemize}
    \item \textbf{Caption-Image retrieval} This task is comprised of two sub-tasks: ranking a large collection of images by their relevance for some given query text (Image Retrieval) and ranking captions by their relevance for some given query image (Caption Retrieval). Both tasks are evaluated on data from the COCO dataset \citep{COCO}. Each image is represented by a pretrained, 2048-dimensional embedding produced by a ResNet-101 \citep{he2016deep}.
\end{itemize}

The probing tasks are designed to evaluate what linguistic properties are encoded in a sentence representation. All tasks are binary or multi-class classification. We give a brief overview of each task below (see \citealt{probing} for more details):

\begin{itemize}
    \item \textbf{Sentence length (SentLen)}: A multi-class classification task where a model is trained to predict the length of a given input sentence, which is binned into six possible length ranges.
    \item \textbf{Word content (WC)}: A multi-class classification task where, given 1000 words as targets, the goal is to predict which of the target words appears in a given input sentence. Each sentence contains a single target word, and the word occurs exactly once in the sentence.
    \item \textbf{Tree depth (TreeDepth)}: A multi-class classification task where the goal is to predict the maximum depth (with values ranging from 5 to 12) of a given input sentence's syntactic tree.
    \item \textbf{Bigram Shift (BShift)}: A multi-class classification task where the goal is to predict whether two consecutive tokens within a given sentence have been inverted.
    \item \textbf{Top Constituents (TopConst)}: A multi-class classification task where the goal is to predict the top constituents (from a choice of 19) immediately below the sentence (S) node of the sentence's syntactic tree.
    \item \textbf{Tense}: A binary classification task where the goal is to predict the tense (past or present) of the main verb in a sentence.
    \item \textbf{Subject number (SubjNum)}: A binary classification task where the goal is to predict the number (singular or plural) of the subject of the main clause.
    \item \textbf{Object number (ObjNum)}: A binary classification task, analogous to SubjNum, where the goal is to predict the number (singular or plural) of the direct object of the main clause.
    \item \textbf{Semantic odd man out (SOMO)}: A binary classification task where the goal is to predict whether a sentence has had a single randomly picked noun or verb replaced with another word with the same part-of-speech.
    \item \textbf{Coordinate inversion (CoordInv)}: A binary classification task where the goal is to predict whether the order of two coordinate clauses in a sentence has been inverted.
\end{itemize}

\subsection{Computing an average score} \label{senteval-average}

In our paper, we present averaged downstream and probing scores. Computing averaged probing scores was straightforward; each of the ten probing tasks reports a simple accuracy, which we averaged. To compute an averaged downstream score, we do the following:

\begin{itemize}
    \item If a task reports Spearman correlation (i.e. SICK-R, STS-B), we use this score when computing the average downstream task score. If the task reports a mean Spearman correlation for multiple subtasks (i.e. STS12, STS13, STS14, STS15, STS16), we use this score.
    \item If a task reports both an accuracy and an F1-score (i.e. MRPC), we use the average of these two scores.
    \item For the Caption-Image Retrieval task, we report the average of the Recall@K, where \(K \in \{1, 5, 10\}\) for the Image and Caption retrieval tasks (a total of six scores). This is the default behaviour of SentEval.
    \item Otherwise, we use the reported accuracy.
\end{itemize}

\end{document}